\documentclass[review]{elsarticle}
 \usepackage{fullpage}
 \usepackage{amsfonts} 
 \usepackage{amsmath}
 \usepackage{pifont}
 \usepackage{natbib}
 \usepackage{geometry}
 \usepackage{graphicx}
 \usepackage{hyperref}
 
\RequirePackage{lineno}
\usepackage{subfig}
\usepackage{amssymb}
\usepackage{multirow}
\usepackage{subfig}
\usepackage{setspace}
\usepackage{url}
\usepackage{color}

\newtheorem{thm}{Theorem}

\newtheorem{cor}{Corollary}

\newtheorem{defi}{Definition}
\newtheorem{assu}{Assumption}
\newproof{proof}{Proof}
\begin{document}

\begin{frontmatter}
\title{Generalized Canonical Correlation Analysis for Classification}
\author[jhuams]{Cencheng Shen}
\ead{cshen6@jhu.edu}
\author[jhuece]{Ming Sun}
\ead{msun8@jhu.edu}
\author[jhuams]{Minh Tang}
\ead{mtang10@jhu.edu}
\author[jhuams]{Carey E. Priebe\corref{cor1}}
\ead{cep@jhu.edu}

\address[jhuams]{Department of Applied Mathematics and Statistics, Johns Hopkins University, Baltimore, MD 21218}
\address[jhuece]{Department of Electrical and Computer Engineering, Johns Hopkins University, Baltimore, MD 21218}

\cortext[cor1]{Corresponding author}

\begin{abstract} 
For multiple multivariate datasets, we derive conditions under which Generalized Canonical Correlation Analysis (GCCA) improves classification performance of the projected datasets, compared to standard Canonical Correlation Analysis (CCA) using only two data sets. We illustrate our theoretical results with simulations and a real data experiment.
\end{abstract}

\end{frontmatter}

\section{Introduction}

With the advent of big data acquisition technology, collected datasets have grown faster than our understanding of how to make optimal use of them. It is common to find collections/measurements of related objects, such as the same article in different languages, similar talks given by different presenters, similar weather patterns in different years, etc. It remains to determine how much the available big data helps us in statistical analysis; simply throwing every collected dataset into the mix may not yield an optimal output. Thus it is natural and important to understand theoretically when and how additional datasets improve the performance of various statistical analysis tasks such as regression, clustering, classification, etc. This is our motivation to explore the following classification problem.

Let $(X,Y) \sim F_{XY}$ be an $\mathbb{R}^{m} \times \{1,\ldots,K\}$ random pair, where $X$ is the feature vector and $Y$ is the class label. In statistical pattern recognition (see, e.g., \cite{DGLPatternRecognitionBook}, \cite{DHSPatternRecognitionBook}) one seeks a classifier $g: \mathbb{R}^{m} \rightarrow \{1,\ldots,K\}$ such that the probability of misclassification $L(g)=P\{g(X)\neq Y\}$ is acceptably small. Because modern datasets are often multi-dimensional, the feature vector $X$ is assumed to be a multivariate random variable of dimension $m$ and it is often beneficial to carry out the classification in some lower dimension $d$ ($1 \leq d < m$) as $m$ is usually large. Therefore dimension reduction is applied to first embed $X$ from $\mathbb{R}^{m}$ to $\mathbb{R}^{d}$, prior to subsequent classification. 

Herein we consider only linear projections, which are commonly used and are the foundation for many nonlinear methods. We denote a linear projection from $\mathbb{R}^{m}$ to $\mathbb{R}^{d}$ by an $m \times d$ matrix $A$; then $A^{'}X$ (the $'$ sign denotes transpose) is the projected feature vector in $\mathbb{R}^{d}$. It follows that the classification error for a given classifier $g$ (whose domain is $\mathbb{R}^{d}$ from now on) is $L_{A} = P\{g(A^{'}X) \neq Y\}$.

Given a distribution $F_{XY}$, a classifier $g$, and a non-empty set of linear projections $\mathcal{A}$, we define an optimal projection $A^{*} \in \arg\min_{A\in \mathcal{A}} \{ L_{A} \} $ and denote the corresponding minimum error as $L_{A^{*}}$. The set $\mathcal{A}$ and the existence of $A^{*}$ are discussed in Section~\ref{preliminaries} and Assumption~\ref{assum}. Roughly speaking, $L_{A^{*}}$ is the minimum error one can hope to achieve by choosing $A$ cleverly among linear projections.

Assuming that the classifier $g$ is specified, the crucial step is to choose the dimension reduction method. If we have only $X$ available as the feature vector, then PCA (Principal Component Analysis) \cite{JolliffePCABook} is a natural choice, which is applied for classification in \cite{YangYangPCALDA2003}. On the other hand, if there is an auxiliary feature $Z_{1}$ of dimension $m_{1}$ available, that is, $(X, Z_{1},Y) \sim F_{XZ_{1}Y}$ on $\mathbb{R}^{m} \times \mathbb{R}^{m_{1}} \times \{1,\ldots,K\}$, then CCA (Canonical Correlation Analysis) \cite{HotellingCCA1936} is applicable on the pair $(X,Z_{1})$ to derive the projection $A$, which is used in \cite{CCAOverview}. In general, if there are $S$ auxiliary features $\{Z_{s} \in \mathbb{R}^{m_{s}}, s=1,\ldots,S \}$ (we always assume $1 \leq d \leq \min{\{m,m_{1},\ldots,m_{S}\}}$), then GCCA (Generalized Canonical Correlation Analysis) \cite{Kettenring1971CCA} is applicable on $(X, Z_{1},\cdots,Z_{S})$ to derive $A$ based on $X$ and the auxiliary features $\{ Z_{s} \}$.

Note that our classification task remains the same, so that at the classification step we observe only $X$ but not $\{Z_{s}\}$; and so by ``GCCA/CCA is applicable'' we mean ``GCCA/CCA can be used to derive the projection matrix $A$ for use in the classifier $g(A^{'}X)$''. Furthermore, although CCA is a special case of GCCA, for clarity purposes we shall assume that GCCA uses at least two auxiliary features whenever GCCA is compared to CCA. If we consider those auxiliary features as extra datasets available for use, GCCA can make use of additional datasets compared to CCA, but we do not know whether these additional datasets will allow GCCA to outperform CCA. At this moment, we should also point out that another popular approach combines GCCA/CCA into the supervised learning step explicitly as a classification rule \cite{HastieBujaTibshirani1995}, \cite{SunJiYeCCA2011}, \cite{TenenhausTenenhausRGCCA2011}, which is empirically more suitable if classification is the only purpose; while in our setting we first apply GCCA/CCA to project the data, followed by the supervised learning step based on the projected data and known labels, which is a more general and more classical view in exploring given data and can be followed by other inference tasks such as testing, clustering, classification, etc. These two approaches are not in conflict with each other: one may first apply GCCA/CCA to project the data without the labels, followed by classification using supervised CCA (which in fact is equivalent to linear discriminant analysis in the two-class case \cite{HastieBujaTibshirani1995}). 

The above setting leads to the following questions. Does GCCA perform better than CCA in classification when using additional auxiliary features? From an application point of view, do additional datasets help in the later classification task, and what type of datasets should be included as auxiliary features in deriving the projection? It turns out the answer is not simple. We consider these questions theoretically, by deriving conditions on the auxiliary features that imply the superiority of GCCA. Let us say the joint feature $(X,Z_{1},\cdots,Z_{S}) \sim F_{S+1}$, and a projection matrix $A$ derived from GCCA/CCA using $X$ and $s$ auxiliary features is denoted by $A_{s+1}$. Our main objective is to derive sufficient conditions on $F_{3}$ such that if $\max{\{L_{A_{2}}\}}=L_{A^{*}}$, then $L_{A_{3}}=L_{A^{*}}$, as well as sufficient conditions such that $L_{A^{*}} = L_{A_{3}} < \min{\{L_{A_{2}}\}}$; and their generalizations to $F_{S+1}$ with arbitrary $s \geq 2$. (Note that when there are two auxiliary features, $A_{2}$ may come from applying CCA to either $(X,Z_{1})$ or $(X,Z_{2})$; hence the `max' and `min'.) Equivalently, the objective is to demonstrate that additional datasets can be useful for the classification task when conditions are satisfied. The necessary prerequisites are discussed in Section~\ref{preliminaries}. The sufficient conditions and the following theorems are shown in Section~\ref{results}. Some discussions are offered in Section~\ref{discuss} to relate the results to practical scenarios such as high-dimensional data and functional data, in addition to the classical multivariate setting the theorems are based on. Our theoretical results are illustrated via simulations, as well as a real data experiment on Wikipedia documents, in Section~\ref{numer}. All proofs are put into Section~\ref{append}, including brief comments to elaborate on the sufficient conditions. 

\section{Preliminaries}
\label{preliminaries}
Given two auxiliary features $Z_{1}$ and $Z_{2}$, the joint distribution of $(X, Z_{1}, Z_{2})$ is denoted by $F_{3} \in \Omega_{3}$, where $\Omega_{3}$ is a family of multivariate distributions on $\mathbb{R}^{(m+m_{1}+m_{2})}$. The overall covariance matrix of $F_{3}$ is denoted by 
\[ 
\Sigma_{F_{3}}= \left [ \begin{array}{ccc} \Sigma_{X} &   \Sigma_{XZ_{1}} &   \Sigma_{XZ_{2}}  \\
\Sigma_{XZ_{1}}^{'} & \Sigma_{Z_{1}} &   \Sigma_{Z_{1}Z_{2}} \\
\Sigma_{XZ_{2}}^{'} & \Sigma_{Z_{1}Z_{2}}^{'} &   \Sigma_{Z_{2}} 
\end{array}
\right ]   \in \mathbb{R}^{(m+m_{1}+m_{2}) \times (m+m_{1}+m_{2})}.
\]
The overall covariance matrix, along with the individual $\Sigma_{X}$, $\Sigma_{Z_{1}}$ and $\Sigma_{Z_{2}}$, are all assumed finite and positive semi-definite with rank no less than $d$.

We can consider GCCA/CCA either with the population covariances or with the sample covariances. For our theoretical analysis we consider the population covariances directly, while in the numerical section we use the sample covariances, which are asymptotically equivalent in the classical multivariate setting under standard regularity conditions \cite{AndersonBook}. 

Identifying the CCA projection $A_{2}=A_{2}(X,Z_{1})$ can be approached as the problem of finding two sets of unit-length canonical vectors $\{a_{i}\}$ and $\{b_{i}\}$ to maximize the correlation between $a_{i}^{'}X$ and $b_{i}^{'}Z_{1}$ for each $i =1,\ldots,d$. (The size of $a_{i}$ is $m \times 1$ and the size of $b_{i}$ is $m_{1} \times 1$.) That is, we wish to identity
\begin{equation}
\label{ccaCond}
\arg\max_{a_{i},b_{i}} \rho_{\{a_{i}^{'}X,b_{i}^{'}Z_{1}\}}=\frac{a_{i}^{'}\Sigma_{XZ_{1}}b_{i}}{\sqrt{a_{i}^{'}\Sigma_{X}a_{i}}\sqrt{b_{i}^{'}\Sigma_{Z_{1}}b_{i}}},
\end{equation}
\begin{center}subject to the \textit{uncorrelated constraints} \end{center}
\begin{align*}
\rho_{\{a_{i}^{'}X,a_{j}^{'}X\}} = \frac{a_{i}^{'}\Sigma_{X}a_{j}}{\sqrt{a_{i}^{'}\Sigma_{X}a_{i}}\sqrt{a_{j}^{'}\Sigma_{X}a_{j}}}=0 \mbox{\ and \ }\rho_{\{b_{i}^{'}Z_{1},b_{j}^{'}Z_{1}\}} = \frac{b_{i}^{'}\Sigma_{Z_{1}}b_{j}}{\sqrt{b_{i}^{'}\Sigma_{Z_{1}}b_{i}}\sqrt{b_{j}^{'}\Sigma_{Z_{1}}b_{j}}}=0, \forall j <i.
\end{align*}
Then the $m \times d$ matrix $A_{2}=[a_{1},\ldots,a_{d}]$ is the CCA projection matrix for $X$, and $A_{2}^{'}X \in \mathbb{R}^{d}$ is the projected feature vector. Alternatively, a different $A_{2}=A_{2}(X,Z_{2})$ can be identified. Note that the arguments to $A_{2}$ -- $(X,Z_{1})$ or $(X,Z_{2})$ -- represent the choice of auxiliary features, and will be suppressed if the choice is clear or irrelevant in the context. 

To identify the GCCA projection $A_{3}$ based on $(X,Z_{1},Z_{2})$, we are looking for three sets of unit-length canonical vectors $\{a_{i}\}$, $\{b_{i}\}$ and $\{c_{i}\}$ as follows:
\begin{equation}
\label{gccaCond}
\begin{split}
&\arg\max_{a_{i}, b_{i}, c_{i}	} (\rho_{\{a_{i}^{'}X,b_{i}^{'}Z_{1}\}}^{r}+\rho_{\{b_{i}^{'}Z_{1},c_{i}^{'}Z_{2}\}}^{r}+\rho_{\{a_{i}^{'}X,c_{i}^{'}Z_{2}\}}^{r}) \\
\mbox{ subject to } & \rho_{\{a_{i}^{'}X,a_{j}^{'}X\}}=\rho_{\{b_{i}^{'}Z_{1},b_{j}^{'}Z_{1}\}}=\rho_{\{c_{i}^{'}Z_{2},c_{j}^{'}Z_{2}\}}=0, \ \forall j <i,
\end{split}
\end{equation}
where the exponent $r$ in the GCCA formulation~(\ref{gccaCond}) indicates the specific GCCA criterion. A common practice is to set $r=1$ or $2$, which maximizes either the sum of correlations or the sum of squared correlations \cite{Kettenring1971CCA}. Then $A_{3}=[a_{1},\ldots,a_{d}]$ is the desired GCCA projection. In general, given $F_{S+1}$ we can derive the GCCA projection $A_{s+1}$ for any $1 \leq s \leq S$, and CCA is merely a special case for $s=1$. Because our results are shown to hold for any $r \geq 1$, we implicitly take $r=1$ unless mentioned otherwise. 

Given $\Sigma_{X}$, we shall call an $m \times d$ matrix $A=[a_{1},\ldots,a_{d}]$ a ``potential" GCCA projection if and only if its columns $\{a_{i}\}$ are of unit-length and satisfy the uncorrelated constraints. The set containing all potential GCCA projections is denoted by $\mathcal{A}=\{A | \ \rho_{\{a_{i}^{'}X,a_{j}^{'}X\}}=0 \ \forall i \neq j \mbox{ and } \|a_{i}\|=1 \ \forall i\}$. As a different choice of auxiliary features yields a different projection, we denote the set containing the GCCA projections $A_{3}$ by $\mathcal{A}_{3}$ and the set containing all CCA projections $A_{2}$ by $\mathcal{A}_{2}$, as well as the set $\mathcal{A}_{s+1}$ in general. Clearly the elements of $\mathcal{A}_{s+1}$ as well as $\mathcal{A}$ depend on $\Sigma_{X}$. Note that the PCA projection is also an element of $\mathcal{A}$, but this is not of our concern in this paper. An important special case: $\mathcal{A}$ represents the Stiefel manifold \cite{ChikuseBook} (containing all orthogonal projections onto dimension $d$ linear subspaces) when $\Sigma_{X}$ is a multiple of the identity. 

Note that the original GCCA/CCA algorithm does not require the norm of $a_{i}$ to be the same for all $i$. We choose them to be unit-length consistently in order to avoid scaling issues in the classification step (alternatively, it is a common practice to set $a_{i}^{'}\Sigma_{X}a_{i}=1$ for all $i$, which is equivalent for our purposes). Also note that the choice of the GCCA/CCA projections can be arbitrary. For example, let $\Sigma_{X}$ and $\Sigma_{Z_{1}}$ be identity matrices and all the singular values of $\Sigma_{XZ_{1}}$ be the same; then $A_{2}(X,Z_{1})$ can be chosen arbitrarily in the Stiefel manifold $\mathcal{V}_{d,m}$. In this case $A_{2}$ has $md-\frac{d^2+d}{2}$ degrees of freedom, where $md$ comes from the dimension freedom by repeating singular values and $\frac{d^2+d}{2}$ comes from the unit-length requirement and uncorrelated constraints. But if $\Sigma_{XZ_{1}}$ does not have repeating singular values, $A_{2}$ represents a fixed subspace and has $\frac{d^2-d}{2}$ degrees of freedom, which is implied by the fact that two $m \times d$ matrices $A$ and $B$ represent the same subspace if and only if $AA^{'}=BB^{'}$. The same phenomenon applies for any GCCA projection $A_{s+1}$.

Returning to the classification problem: given a classifier $g: \mathbb{R}^{d} \rightarrow \{1,\ldots,K\}$ for the low-dimensional feature vector $A^{'}X$, the error $L_{A}$ may differ for different $A \in \mathcal{A}$. Clearly $\mathcal{A}$ is compact for finite $\Sigma_{X}$ and $\{L_{A}|A \in \mathcal{A}\}$ is bounded between $[0,1]$, but an optimal low-dimensional projection (with respect to the classification error) is not guaranteed to exist. We make the following assumption to avoid non-existence:

\begin{assu}
\label{assum}
Given a classifier $g$, we assume for the theory in the sequel that an optimal projection $A^{*}=\arg\min_{A\in \mathcal{A}} \{ L_{A} \}$ exists for any finite $\Sigma_{X}$ of rank at least $d$. 
\end{assu}
For example, if the class-conditional distributions $F_{X|Y=k}$ admit probability density functions $f_{X|Y=k}$ for $k=1,\ldots,K$, then the assumption always holds. (In this case $L_{A}$ is continuous with respect to $A$, and thus $\{L_{A}|A \in \mathcal{A}\}$ is compact and admits a minimum.)

By this assumption, the minimum error $L_{A^{*}}$ always exists and it follows that $L_{A_{s+1}} \geq L_{A^{*}}$ always holds for any $s$. Note that the optimal projection $A^{*}$ need not be unique, since the existence suffices for our purposes. Now we are able to define the notion that GCCA improves CCA using $L_{A^{*}}$.

\begin{defi}
\label{improve}
Assuming the existence of $A^{*}$, we say GCCA improves CCA within a family of distributions $\Omega_{3}$ if and only if $\{F_{3} \in \Omega_{3}|L_{A_{2}}=L_{A^{*}}, \ \forall A_{2} \in \mathcal{A}_{2}\} \subset \{F_{3} \in \Omega_{3}|L_{A_{3}}=L_{A^{*}}, \ \forall A_{3} \in \mathcal{A}_{3}\}$.

In general, we say the set of GCCA projections $\mathcal{A}_{s+1}$ improves the set of GCCA projections $\mathcal{A}_{t+1}$ within $\Omega_{S+1}$ ($1 \leq s,t \leq S$) if and only if $\{F_{S+1} \in \Omega_{S+1}|L_{A_{t+1}}=L_{A^{*}}, \ \forall A_{t+1} \in \mathcal{A}_{t+1}\} \subset \{F_{S+1} \in \Omega_{S+1}|L_{A_{s+1}}=L_{A^{*}}, \ \forall A_{s+1} \in \mathcal{A}_{s+1}\}$. (Here the notation ``$\subset$" indicates proper subset.)
\end{defi}
Put in words, suppose GCCA improves CCA within $\Omega_{3}$. Then the optimality of the CCA projections implies the optimality of the GCCA projection, and there exists $F_{3}$ such that the GCCA projection is optimal while at least one of the CCA projections is not. Such improvement implies that additional datasets should be used, though it is not equivalent to $L_{A_{3}} \leq L_{A_{2}}$. 

If $\Omega_{3}$ includes every possible multivariate distribution, then GCCA fails to improve CCA. For example, if $Z_{1}$ and $Z_{2}$ are both positively correlated to $X$ but $Z_{1}$ and $Z_{2}$ are negatively correlated, then it might happen that $A_{2}$ is optimal while $A_{3}$ is not. Hence it is not always a good idea to incorporate additional auxiliary features, and we shall look for a family $\Omega_{3}$ imposing certain relationships among $X$ and $\{Z_{s}\}$ such that GCCA is guaranteed to improve CCA. 

First, we transform $X$ by centering and whitening, so that the population mean is zero and the population covariance matrix becomes the identity matrix. Then $\mathcal{A}$ consists of orthogonal projections onto dimension $d$ linear subspaces, and there exists an orthogonal matrix such that the feature vector can be rotated to guarantee $A^{*}$ is equivalent to the subspace $\mathbb{R}^{d}$ spanned by the first $d$ coordinate axes. We denote the transformed random variable by $\tilde{X}=H_{X}(X-E(X))$, where $E(X)$ is the expectation for centering and $H_{X}$ is a non-singular $m \times m$ matrix for whitening and rotation. Since the optimal projection for $\tilde{X}$ is spanned by the first $d$ coordinate axes, the form of $\tilde{X}$ based on the class label $Y=\{1,\ldots,K\}$ can be expressed as:
\begin{equation}
\label{X}
\tilde{X} = H_{X}(X-E(X)) \stackrel{law}{=} \left[ {\begin{array}{cc}
 U_{1}\bold{1}_{1}+U_{2}\bold{1}_{2}+\cdots+U_{K}\bold{1}_{K}  \\
 \\
 W \\ \end{array} } \right],
\end{equation}
where $\bold{1}_{k}$ is the class label indicator taking value $k$ with probability $p_{k}$ and $\sum_{k=1}^{K}p_{k}=1$, each $U_{k} \in \mathbb{R}^{d}$ is the marginal distribution of $\tilde{X}$ under class $k$, and $W \in \mathbb{R}^{m-d}$ is the ``irrelevant" marginal of $\tilde{X}$. By the above transformation it holds that $E(W)=0_{(m-d) \times 1}$ and $E(WW^{'})=I_{(m-d) \times (m-d)}$, where $I$ denotes the identity matrix. Clearly $H_{X}$ always exists, and there are multiple choices for $H_{X}$ if $A^{*}$ is not unique. Now we impose our conditions on $F_{S+1}$ and define what we call the similar family.

\section{Main Results}
\label{results}
\begin{defi}
\label{XYZ}
We say the family of distributions $\Omega_{S+1}^{*}$ is \emph{the similar family} if and only if it includes every $F_{S+1}$ such that $(X, Z_{1}, \cdots, Z_{S}) \sim F_{S+1}$ satisfies the following conditions:

Condition (1): For each $A^{*}$, there exists non-singular matrices $H_{X} \in \mathbb{R}^{m \times m}$ and $H_{Z_{s}} \in \mathbb{R}^{m_{s} \times m_{s}}$ for all $s=1,\ldots,S$, such that Equation~(\ref{X}) holds and there exist non-negative scalars $q_{sk}$ with
\begin{equation}
\label{YZZ}
\tilde{Z}_{s} = H_{Z_{s}}(Z_{s}-E(Z_{s})) \stackrel{law}{=} \left[ {\begin{array}{cc}
 q_{s1}U_{1}\bold{1}_{1}+q_{s2}U_{2}\bold{1}_{2}+\cdots+q_{sK}U_{K}\bold{1}_{K}+e_{s}  \\
 \\
 W_{s} \\ \end{array} } \right],
\end{equation}
where $e_{s}$ represents independent noise and $W_{s} \in \mathbb{R}^{m_{s}-d}$. Note that unlike $H_{X}$, $H_{Z_{s}}$ need only be non-singular and $Z_{s}$ are not necessarily whitened and rotated.

Condition (2): $E(U_{k}U_{k}^{'})=I$, and $U_{k}$ is uncorrelated with $W$ and $W_{s}$, for all $k=1,\ldots,K$ and $s=1,\ldots,S$. 

Condition (3): $\sigma_{1}(E(W_{s}W_{t}^{'})) \leq \sigma_{1}(E(WW_{s}^{'}))\sigma_{1}(E(WW_{t}^{'}))$ for all $1 \leq s \neq t \leq S$, where we denote $\sigma_{i}(\Sigma)$ as the $i$th largest singular value for any matrix $\Sigma$ henceforth.

Condition (4): $(q_{sk_{1}}-q_{sk_{2}})(q_{tk_{1}}-q_{tk_{2}}) > 0$ for all $1 \leq s < t \leq S$ and $k_{1},k_{2}=1,\ldots,K$; namely the ordering of coefficients $q_{sk}$ is consistent throughout $Z_{s}$. 
\end{defi}

The purpose of condition (1) is to guarantee that the marginal distribution restricted to $A^{*}$ of every transformed auxiliary feature under each class is a scalar multiple of the corresponding marginal of $\tilde{X}$ plus error. The possible non-uniqueness of $A^{*}$ is (mostly) avoided by requiring (1) to hold for any $A^{*}$, though the transformation matrices and respective scalars probably differ under different $A^{*}$. Condition (2) is to simplify the analysis, without which the proof is much more complex. Given conditions (1) and (2), conditions (3) and (4) are technical conditions used in the proof, implying certain relationships among features. Interpreted by words, condition (3) implies the ``noisy'' dimensions (where $W$ and $W_{s}$ live in) among the auxiliary features should be less related, while condition (4) implies the ``signal'' dimensions (where $U_{k}$ lives) among the auxiliary features should be more related. In this case GCCA is more likely to extract information from the ``signal'' dimensions, for which utilizing additional datasets is likely to improve the classification error. As we will see in the numerical experiments, this interpretation is useful for judging qualitatively whether additional datasets should be included, even if $A^{*}$ is unknown or condition (2) is not satisfied. And we will provide additional comments at the end of the proof section to discuss the magnitude of $q_{sk}$ and its potential impact on the sufficient conditions and model selection.

\begin{thm}
\label{main}
GCCA improves CCA in the similar family $\Omega_{3}^{*}$.
\end{thm}

Therefore it is beneficial to use the GCCA projection $A_{3}$ within the similar family $\Omega_{3}^{*}$, whose conditions are sufficient but not necessary for GCCA to improve CCA. Equivalently, deriving the projection using additional datasets helps the classification task when the sufficient conditions are satisfied.

Furthermore, the similar family can be decomposed into three disjoint subsets as follows: $\Omega_{3}^{*} = \{F_{3}\in \Omega_{3}^{*}| \max{\{L_{A_{2}}\}}=L_{A_{3}}=L_{A^{*}}\} \cup \{F_{3}\in \Omega_{3}^{*}| \max{\{L_{A_{2}}\}} > L_{A_{3}}=L_{A^{*}}\} \cup \{F_{3}\in \Omega_{3}^{*}| \max{\{L_{A_{2}}\}} > L_{A^{*}} \textup{ and } L_{A_{3}}>L_{A^{*}}\}$, with all the subsets shown to be non-empty and proper in the proof (we can also replace all the `max' by `min'). Specifically, if the optimal $A^{*}$ is known (which may be difficult in practice), then one can check which subset a given $F_{3} \in \Omega_{3}^{*}$ belongs to according to Inequality~(\ref{cond1}) and Inequality~(\ref{cond2}) in the proof below. When the distribution lies in the first or the second subset above, the GCCA projection performs no worse than the CCA projections, and adding a ``qualified'' additional dataset yields better classification result.

It is natural to consider a generalization to $\Omega_{S+1}^{*}$ because there may be many additional datasets satisfying the conditions. Indeed we have an easy generalization of the above theorem.

\begin{cor}
\label{main2}
For any $S \geq S^{'} \geq 2$, the set of GCCA projections $\mathcal{A}_{S^{'}+1}$ improves the set of CCA projections $\mathcal{A}_{2}$ in the similar family $\Omega_{S+1}^{*}$.
\end{cor}

Under a simplified setting, we can also show that the set of GCCA projections continue to improve when additional auxiliary features are included in deriving the projections. This means in the context of the similar family, additional datasets will always improve the performance in the classification task.

\begin{cor}
\label{main3}
Let us replace condition (4) by a simplifying condition (4'): $W_{s}=W_{t}$ and $q_{sk}=q_{tk}$ for all $1 \leq s,t \leq S$. Namely the auxiliary features follow the same distribution for $s=1,\ldots,S$.

Then for any $S \geq S^{'} \geq 2$, the set of GCCA projections $\mathcal{A}_{S^{'}+1}$ always improves the set of GCCA projections $\mathcal{A}_{S^{'}}$ in the similar family $\Omega_{S+1}^{*}$.
\end{cor}

\section{Discussions}
\label{discuss}
Since our analysis is carried out on the population covariance instead of the sample covariance, our results so far rely on the fact that the sample covariance converges to the population covariance as dimension reduction methods including GCCA/CCA are mostly carried out on the sample data. Let us provide some justifications for the high-dimensional data case, where the dimension $m$ is large when compared to the number of training observations $n'$ such that the covariance convergence is not guaranteed.

For high-dimensional data, if the sample covariance is still close to the true covariance with high probability as discussed in \cite{VershyninClosenessCovariance2012} and \cite{VershyninCovariance2013}, then our results still apply and GCCA improves CCA in the similar family with high probability. Otherwise our conditions in Definition~\ref{XYZ} cannot be directly used to justify the GCCA/CCA behavior on sample covariances of high-dimensional data. However, one may heuristically claim that if GCCA is better than CCA in the population model for the classification task, then GCCA is expected to be better than CCA for the sample data: Since the classification error is actually a function of the data, if $L_{A_{3}} < L_{A_{2}}$ for $A_{2}$ and $A_{3}$ derived from the population model, then at a suitable level of $n'/m$ we can have $Prob\{L_{A_{3}} < L_{A_{2}}\} > 0.5$ for $A_{2}$ and $A_{3}$ derived from the sample data, because this probability converges to $1$ in the classical multivariate setting where $n'/m \rightarrow \infty$. (A point of interest is to derive the minimum level $n'/m$, which may depend on the classifier we use. For our simulations on the synthetic data generated within the similar family, it seems the minimum level is no larger than 1 in order for GCCA to be better than CCA.)

In practice one rarely applies CCA directly on data of very high dimension with $m > n$. Often one opts to use kernel CCA \cite{HardoonKernelCCA2007}, \cite{LaffertyCCA2012}, sparse CCA \cite{HardoonSparseCCA2011}, \cite{WittenTibshiraniHastie2009} or functional CCA \cite{HwangJungTakaneCCA2012}, \cite{HeMullerWangCCA2003} to deal with noisy high-dimensional data, assuming that the data intrinsically lives in some low-dimensional linear subspace. For example, instead of working on $(X,Y) \in \mathbb{R}^{m}$ where $m$ is very large, kernel/functional CCA works on $(f(X),g(Y))$ by assuming appropriate $f$ and $g$ exist for nonlinear/functional data. But the analysis of sparse/functional CCA will be quite different and difficult when penalty terms are introduced in the constraints, which requires numerical methods to solve and gives different GCCA/CCA transformations that cannot be efficiently expressed in matrix notation.

Another aspect worth noting is that a similar conclusion may be reached for clustering. This is because GCCA makes it easier to find the optimal subspace than CCA under the same conditions, as long as one is able to define an optimal subspace $A^{*}$ in terms of some clustering algorithm with respect to a specific performance index. However, we do not pursue this direction here because it is more challenging to evaluate clustering performance than classification performance. 

Furthermore, since GCCA/CCA does not make use of label information in the dimension reduction step, it is natural to compare with some existing algorithms such as p-LDA (penalized linear discriminant analysis) \cite{HastieBujaTibshirani1995}, \cite{WittenTibshirani2011} and $\ell 1$-SVM ($1$-norm support vector machine) \cite{ZhuRossetHastieTibshirani2003}, \cite{fm2004}, which make use of labels and may work for data of high/unknown dimensions. Even though we will include their classification results in the numerical section for benchmark purposes, our target is not to find the best method for a given dataset. In addition to being more appropriate for an exploratory task, there are other reasons that applying unsupervised dimension reduction methods first is more favorable than doing supervised dimension reduction directly, e.g., it is easier and faster to use unsupervised dimension reduction for real data, it may be slow and difficult to choose a suitable penalty term in p-LDA, the data before dimension reduction may not have access to the labels or may be different from the data on which we perform classification as in the transfer learning task \cite{PanYang2010}, etc. 

At last, the choice of projection dimension $d$ is crucial for the classification (or any inference) performance, especially when working with real data of unknown true dimension. There are a number of papers on dimension choice for projecting a single dataset \cite{ZhuGhodsiAutomaticDimensionSelection2006}, \cite{HoyleAutomaticDimensionSelection2008} but not for multiple correlated datasets, which may be an interesting point to pursue. Still, our results are always valid no matter the choice of $d$, which means GCCA improves CCA for any $d$ when conditions are satisfied. 

\section{Numerical Experiments}
\label{numer}
To investigate the performance of the GCCA/CCA projections in classification, we present both numerical simulations and a real data experiment. We use sample covariances to derive the GCCA projections with the GCCA algorithm implemented according to \cite{TenenhausTenenhausRGCCA2011} (though no covariance matrix regularization is required in our experiments in contrast to their RGCCA algorithm; and we apply Gram-Schmidt to all output vectors in the iteration of the algorithm to enforce the uncorrelated constraints of all the canonical vectors), and the usual LDA as our main classification rule for the following supervised learning. Whenever applicable, we also include p-LDA and $\ell 1$-SVM classification results based on the single dataset to compare with the LDA classification results based on the GCCA/CCA projected dataset. Note that our previous numerical work illustrating GCCA improvement under kNN (k-nearest neighbor) classifier is available in \cite{sptGCCA}. 

\subsection{Numerical Simulations}
We start with four random variables $U_{1}, U_{2}\in \mathbb{R}^3$ and $V_1,V_2\in \mathbb{R}^6$ all independently normally distributed. The parameters are set as follows: $E(U_{1}U_{1}')=E(U_{2}U_{2}')=I_{3 \times 3}$, $E(U_{1})=-E(U_{2})=0.2_{3 \times 1}$, $E(V_1 V'_1) = E(V_2 V'_2) = 0.5I_{6 \times 6}$, $E(V_1)=E(V_2)=0_{6 \times 1}$.

The three random variables $X, Z_{1}, Z_{2}\in\mathbb{R}^9$ are constructed as follows:
\begin{equation}
X \stackrel{law}{=} {U_{1}\bold{1}_1+U_{2}\bold{1}_2 \brack V_1+V_2}, \ \
Z_{1} \stackrel{law}{=} {0.6U_{1}\bold{1}_1+0.4U_{2}\bold{1}_2+e_1 \brack V_1+e_3}, \ \
Z_{2} \stackrel{law}{=} {0.6U_{1}\bold{1}_1+0.4U_{2}\bold{1}_2+e_2 \brack V_2+e_4},
\end{equation} 
where $e_1, e_2\stackrel{\text{i.i.d.}}{\sim} N(0, 0.75I_{3 \times 3})$, $e_3, e_4\stackrel{\text{i.i.d.}}{\sim} N(0, 0.5I_{6 \times 6})$, $\bold{1}_{1}$ and $\bold{1}_{2}$ are class label indicators having equal probability. Using LDA, it is clear that at $d=3$ the ideal optimal projection $A^{*}$ uniquely represents the subspace spanned by the first $d$ coordinate axes. Hence we can fit the joint distribution into Definition~\ref{XYZ} with $d=3$, such that $q_{11}=q_{21}=0.6$, $q_{12}=q_{22}=0.4$, $W=V_{1}+V_{2}$, $W_{1}=V_1+e_3$, $W_{2}=V_2+e_4$, etc. This joint distribution satisfies the required conditions, so it belongs to $\Omega_{3}^{*}$. Further, by checking Inequality~(\ref{cond1}) and Inequality~(\ref{cond2}) in the proof, the joint distribution is actually an element of the subset $\{F_{3}\in \Omega_{3}^{*}| \max{\{L_{A_{2}}\}} > L_{A_{3}}=L_{A^{*}}\} \in \Omega_{3}^{*}$. So we expect GCCA to outperform CCA when projected onto $\mathbb{R}^{3}$. Note that in this case we can explicitly calculate $L^{*}$ for the population model, which is $36.45\%$.

For each Monte Carlo replicate, $n=1500$ observations are generated for each random variable. That is, $\{x^{(1)},\ldots,x^{(1500)}\}$ for $X$, $\{z_{1}^{(1)},\ldots,z_{1}^{(1500)}\}$ for $Z_{1}$ and $\{z_{2}^{(1)},\ldots,z_{2}^{(1500)}\}$ for $Z_{2}$. All data points are used to learn the GCCA/CCA projections respectively for $d=3$. (One may instead derive the projections based on the training data only, which is asymptotically equivalent to deriving the projections from all the available data if the testing data is distributed the same as the training.) Then the first 1000 points generated from $X$ are projected and used to train the classifier; the remaining 500 points are projected and used for classification error testing. The classification error is recorded separately for the CCA projections $A_{2}(X,Z_{1})$ and $A_{2}(X,Z_{2})$ and for the GCCA projections $A_{3}$, using both sum of correlation ($r=1$) and sum of squared correlation ($r=2$) criteria. The above is done for 500 Monte Carlo replications, and we show in Table \ref{table:simu1} the average classification error and the average difference between the derived GCCA/CCA subspace and the optimal subspace for each projection (we use the Hausdorff distance \cite{QZLMetrics2005} for the difference between subspaces). The average GCCA classification error is lower than that of CCA as expected, and is fairly close to the optimal error $L^{*}$. In this case the average errors using the p-LDA and $\ell 1$-SVM are $37.37\%$ and $36.50\%$ respectively (the penalty terms are always chosen based on cross-validation for the best performances and benchmark purposes). Note that the standard deviations for the average errors of all the methods are within $0.3\%$, and those for the distance of the subspaces are within $0.002$, which are the same for all the later simulations. Also note that the distances of the subspaces are not expected to be $0$, because the $A^{*}$ we use is the ideal optimal subspace for the population model and different from the optimal subspace for the sample data; but even so, it seems that the classification error is positively correlated to the distance of the subspaces.

To investigate the effect of higher dimension and less sample data, we repeat the same procedure three times, for $m=20$ with $n=1500$, $m=50$ with $n=1500$, and $m=50$ with $n=75$ ($50$ points used for training and the remaining $25$ used for testing). The settings are the same with $d=3$ fixed, e.g., the dimensions of $U_{i}$ stay at $3$ but the dimensions of $V_{i}$ are increased as $m$ increases. The results are shown in Table \ref{table:simu2}, Table \ref{table:simu3} and Table \ref{table:simu4}. A higher dimension or a smaller training size means the sample covariance does a worse job in approximating the population covariance, possibly making the differences between the derived GCCA/CCA subspace and the optimal subspace larger as $m$ increases and/or $n$ decrease; but still GCCA is better than CCA for the classification task in all the tables, reflecting our heuristic argument in the discussion section. This time the average errors using the p-LDA and $\ell 1$-SVM are $39.01\%$ and $39.27\%$ at $m=20$ with $n=1500$, $38.91\%$ and $38.81\%$ at $m=50$ with $n=1500$, and $47.43\%$ and $45.76\%$ at $m=50$ with $n=75$, most of which turn out to be slightly better than using LDA on GCCA projected data throughout these simulations.

We also present another simulation to show that GCCA does not necessarily improve CCA at $m=9$ with $n=1500$, by replacing the auxiliary feature $Z_{2}$ by $Z_{2'} \stackrel{law}{=} {0.6U_{1}\bold{1}_1+0.4U_{2}\bold{1}_2+e_2 \brack V_1+e_4}$. We re-generate all observations and carry out the same simulation steps. Although the auxiliary feature $Z_{2'}$ looks reasonably ``similar'' to $X$ (differing from $Z_{1}$ only by noise), the joint distribution of $(X,Z_{1},Z_{2'})$ does not satisfy condition (3) and GCCA does not improve CCA by checking the covariance structure explicitly. Interpreted by words, $Z_{1}$ and $Z_{2'}$ are too correlated in the ``noisy'' dimensions, hindering GCCA from recognizing the correct ``signal'' dimensions. The average simulated classification errors are shown in Table \ref{table:simu5}. In this case GCCA performs worse than CCA, which demonstrates that simply adding more datasets does not automatically yield a better result.

\begin{table*}[!t]
\renewcommand{\arraystretch}{1.3}
\centering
{\begin{tabular}{|c||c|c|c|c|}
\hline
projections & CCA on $(X,Z_{1})$ & CCA on $(X,Z_{2})$ & GCCA $(r=1)$ & GCCA $(r=2)$  \\
\hline
average error ($L_{A}$) & $42.03\%$ & $41.89\%$  & $37.00\%$ & $38.16\%$ \\
\hline
$\|A-A^{*}\|$  & $1.688$ & $1.591$ & $0.714$ & $0.989$\\
\hline
\end{tabular}
\caption{GCCA Improves CCA in simulation at $m=9, n=1500$}
\label{table:simu1}
}
\end{table*}

\begin{table*}[!t]
\renewcommand{\arraystretch}{1.3}
\centering
{\begin{tabular}{|c||c|c|c|c|}
\hline
projections & CCA on $(X,Z_{1})$ & CCA on $(X,Z_{2})$ & GCCA $(r=1)$ & GCCA $(r=2)$ \\
\hline
average error ($L_{A}$)  & $47.02\%$ & $46.18\%$  & $42.84\%$ & $44.19\%$ \\
\hline
$\|A-A^{*}\|$ & $2.161$ & $2.037$ & $1.364$ & $1.825$ \\
\hline
\end{tabular}
\caption{GCCA Improves CCA in simulation at $m=20, n=1500$}
\label{table:simu2}
}
\end{table*}

\begin{table*}[!t]
\renewcommand{\arraystretch}{1.3}
\centering
{\begin{tabular}{|c||c|c|c|c|}
\hline
projections & CCA on $(X,Z_{1})$ & CCA on $(X,Z_{2})$ & GCCA $(r=1)$ & GCCA $(r=2)$ \\
\hline
average error ($L_{A}$) & $47.58\%$ & $46.02\%$  & $42.41\%$ & $44.31\%$ \\
\hline
$\|A-A^{*}\|$  & $2.197$ & $2.161$ & $1.643$ & $1.895$ \\
\hline
\end{tabular}
\caption{GCCA Improves CCA in simulation at $m=50, n=1500$}
\label{table:simu3}
}
\end{table*}

\begin{table*}[!t]
\renewcommand{\arraystretch}{1.3}
\centering
{\begin{tabular}{|c||c|c|c|c|}
\hline
projections & CCA on $(X,Z_{1})$ & CCA on $(X,Z_{2})$ & GCCA $(r=1)$ & GCCA $(r=2)$ \\
\hline
average error ($L_{A}$) & $51.98\%$ & $51.60\%$  & $45.76\%$ & $49.98\%$ \\
\hline
$\|A-A^{*}\|$  & $2.256$ & $2.236$ & $2.179$ & $2.203$ \\
\hline
\end{tabular}
\caption{GCCA Improves CCA in simulation at $m=50, n=75$}
\label{table:simu4}
}
\end{table*}

\begin{table*}[!t]
\renewcommand{\arraystretch}{1.3}
\centering
{\begin{tabular}{|c||c|c|c|c|c|}
\hline
projections &CCA on $(X,Z_{1})$ & CCA on $(X,Z_{2'})$ & GCCA $(r=1)$ & GCCA $(r=2)$ \\
\hline
average error ($L_{A}$) & $41.34\%$ & $41.33\%$  & $46.86\%$ & $46.90\%$\\
\hline
$\|A-A^{*}\|$  & $1.545$ & $1.537$ & $2.009$ & $2.018$  \\
\hline
\end{tabular}
\caption{GCCA Fails to Improve CCA in simulation}
\label{table:simu5}
}
\end{table*}

\subsection{Wikipedia Documents}
The real data experiment applies GCCA/CCA to text document classification. The dataset is obtained from Wikipedia, an open-source multilingual web-based encyclopedia with millions of articles in more than 280 languages. In Wikipedia each article can be related to others in the same language, or articles in other languages with the same subject. Articles of the same subject in different languages are not necessarily exact translations of one another; it is very likely they are written by different people and their contents might differ significantly.

English articles within a 2-neighborhood of the English article ``Algebraic Geometry" are collected, and the corresponding French articles of those English documents are also collected, which totals $n=1382$ pairs of articles in English and French. Let $a^{e}_1,\ldots, a^{e}_{1382}$ denote the English articles and $a^{f}_1,\ldots, a^{f}_{1382}$ denote the French articles. All articles are manually labeled into $5$ disjoint classes ($1-5$) based on their topics, as shown in Table \ref{tbl:wikitpc}. 

\begin{table*}[!t]
\centering
\begin{tabular}{|c||c|c|c|c|c|}
\hline
topic & category & people & locations & date & math \\ 
\hline
class label & 1 & 2 & 3 & 4 & 5 \\
\hline
article number & 119 & 372 & 270 & 191 & 430 \\
\hline
\end{tabular}
\caption{Wikipedia Dataset Topics}
\label{tbl:wikitpc}
\end{table*}

For the purposes of GCCA/CCA, first we need to embed each article onto the Euclidean space $\mathbb{R}^{m}$ by Multi-dimensional Scaling (MDS) \cite{TorgersonBook1}, \cite{CoxBook}, \cite{BorgBook}. MDS strives to give a Euclidean representation while approximately preserving the dissimilarities of the original data: given an $n \times n$ dissimilarity matrix $\Delta = [\delta_{ij}]$ for $n$ observations with $\delta_{ij}$ being the dissimilarity measure between the $i$th and $j$th observation, MDS generates embeddings $x_{i} \in \mathbb{R}^{m}$ for the $i$th data point to preserve the dissimilarity among the objects pairs, i.e. $||x_i - x_j|| \approx \delta_{ij}$. 

For our work two different types of dissimilarity measures are considered for English and French articles, giving four dissimilarity matrices of dimension $1382 \times 1382$: the graph topology dissimilarity matrix $\bar{\Delta}^{e}, \bar{\Delta}^{f}$ and the text content dissimilarity matrix $\hat{\Delta}^{e}, \hat{\Delta}^{f}$.

For the graph dissimilarities, $\bar{\Delta}^{e}$ and $\bar{\Delta}^{f}$ are constructed based on an undirected graph $G(V, E)$, where $V$ represents the set of vertices of the $1382$ Wikipedia documents, and $E$ is the set of edges connecting those articles. There is an edge between two vertices if they are linked in Wikipedia. Then the entry $\bar{\Delta}^{e}(i,j)$ is calculated from the number of steps on the shortest path from document $i$ to document $j$ in $G$. For the English articles, $\bar{\Delta}^e(i,j) \in \{0, \ldots, 4, 6\}$, where $4$ is the upper bound of the step number with any higher number setting to $6$. For the French articles $\bar{\Delta}^{f}(i,j)$ depends on the French graph connections, so it is possible that $\bar{\Delta}^{f}(i,j) \neq \bar{\Delta}^{e}(i,j)$. At the extreme end, $\bar{\Delta}^{f}(i,j) = \infty$ when $a^f_i$ and $a^f_j$ are not connected, and we set $\bar{\Delta}^{f}(i,j) = 6$ for $\bar{\Delta}^{f}(i,j) > 4$.

For the text dissimilarities, $\hat{\Delta}^{e}$ and $\hat{\Delta}^{f}$ are based on the text processing features for documents $\{a^{e}_{i}\}$ and $\{a^{f}_{i}\}$. Suppose $\mathbf{z}_i, \mathbf{z}_j$ are the feature vectors for the $i$th and $j$th English articles. Then $\hat{\Delta}^{e}(i,j)$ is calculated by the cosine dissimilarity $\hat{\Delta}^{e}(i,j) = 1 - \frac{\mathbf{z}_i \cdot \mathbf{z}_j}{\|\mathbf{z}_i\|_2 \|\mathbf{z}_j\|_2}$. For the experiment we consider the latent semantic indexing (LSI) features \cite{deerwester90}. 

\begin{table*}[!t]
\centering
\begin{tabular}{|c||c|c|}
\hline
 & Graph Topology Dissimilarity & Text Content Dissimilarity \\ 
\hline
English articles $\{a^{e}_i\}$ & $\{\bar{x}^e_i\} (GE)$ & $\{\hat{x}^e_i\} (TE)$ \\
\hline
French articles $\{a^{f}_i\}$ & $\{\bar{x}^f_i\} (GF)$ & $\{\hat{x}^f_i\} (TF)$ \\
\hline
\end{tabular}
\caption{Euclidean Embeddings ($\mathbb{R}^{m}$) for Wikipedia Articles}
\label{tbl:artemb}
\end{table*}

\begin{figure}[htbp]	
  \centering
		\includegraphics[width=1.10\textwidth]{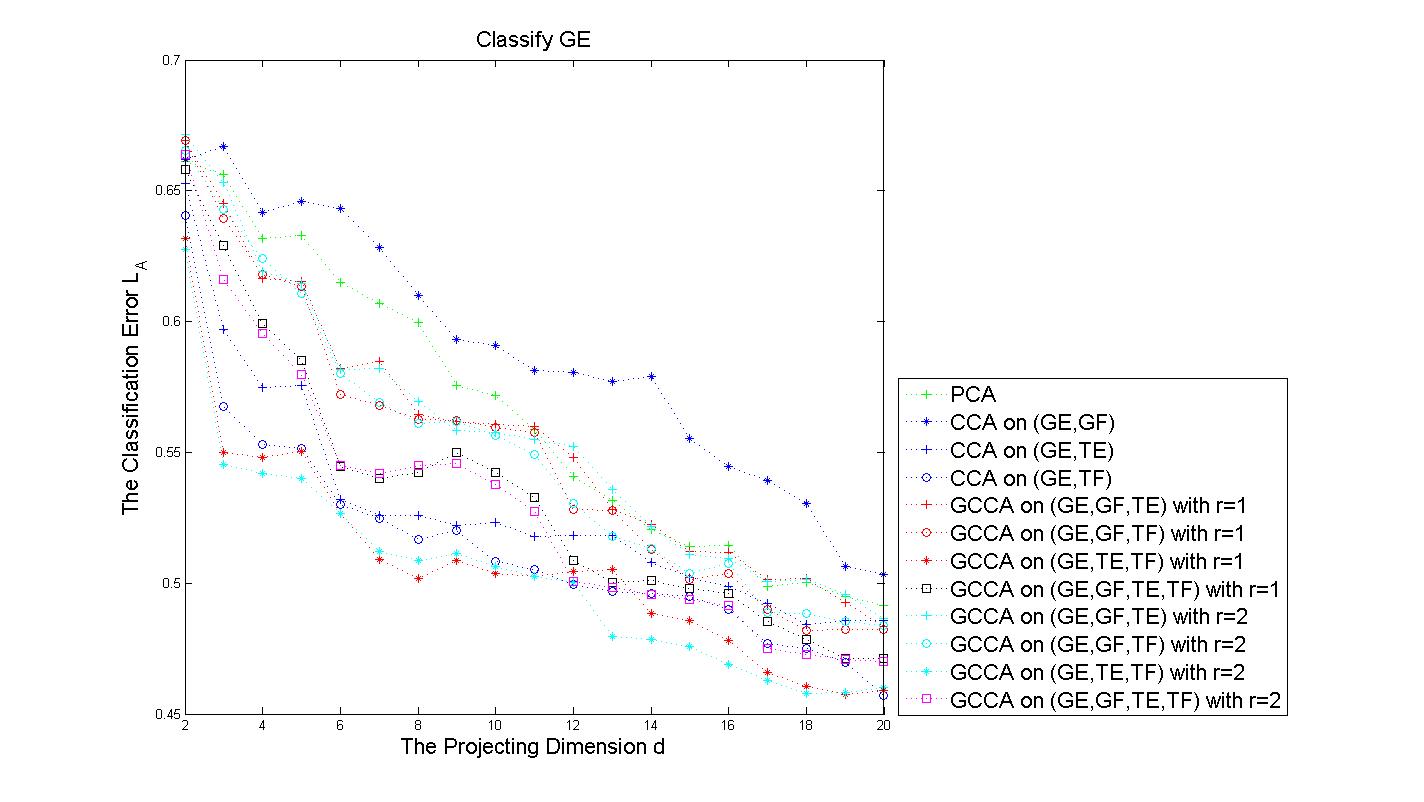}
	\caption{Classification Error for GE}
	\label{fig1}
\end{figure}

\begin{figure}[htbp]
\centering
\subfloat[]{
\includegraphics[width=0.70\textwidth]{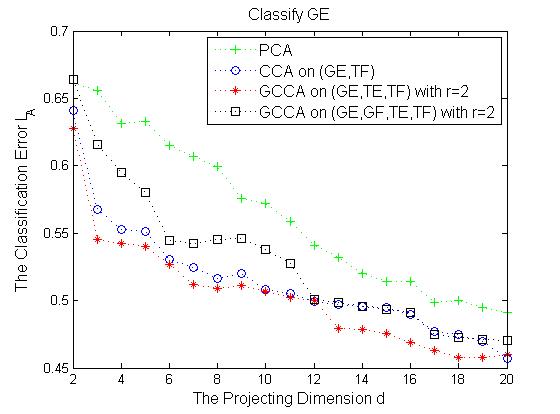}
}
\hfil
\subfloat[]{
\includegraphics[width=0.70\textwidth]{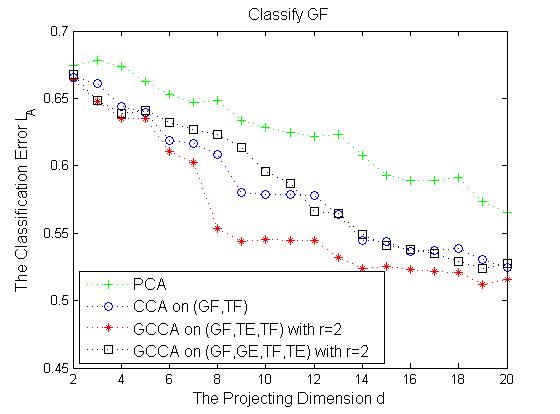}
}
\caption{Classification Error for GE/GF (simplified)}
\label{fig2}
\end{figure}

Once different dissimilarity matrices are constructed, the Euclidean space embeddings with $m=50$ are obtained via MDS. The articles' embeddings are shown in Table \ref{tbl:artemb}. At first, English graph dissimilarity (GE) is the classification target, and others (GF, TE, TF) are treated as auxiliary features: all data points are used to learn the GCCA/CCA projections from $\mathbb{R}^{m}$ to $\mathbb{R}^{d}$ based on GE and a certain choice of auxiliary features, and the data points of GE are projected by the learned projections. Then $600$ observations are randomly picked to train the classifier, with the remaining $782$ documents used for classification error testing. We repeat 500 times to calculate the average classification error, for every possible GCCA/CCA projection and various choice of $d$. The same procedure is repeated with the French graph dissimilarity (GF) being the classification target and the remaining being the auxiliary features. The full results for every possible projection are shown in Figure \ref{fig1} for the classification of GE. For illustration purposes, two simplified plots are shown in Figure~\ref{fig2} for the classification of GE/GF, for which we omit most projections in order to better quantify the effects of increasing $s$ (the number of chosen auxiliary features), i.e., only the best $A_{2}$ and $A_{3}$ are shown. Note that for comparison purposes the PCA projections are also included, and all the classification errors have standard deviations within $0.2\%$. 

Based on Figure~\ref{fig2}, we observe that for most choices of $d$ the best GCCA projection $A_{3}$ admits a lower error than the best CCA projection $A_{2}$, and both of them are better than the PCA projection. The figure also illustrates the last paragraph of our discussion section, i.e., GCCA is expected to be better than CCA no matter the choice of projection dimension. However, it turns out that the GCCA projection $A_{4}$ does not yield the lowest error for classifying the Wikipedia data. This is not a surprise and tells that not all datasets should be included in this example, as one can judge from Figure~\ref{fig1} and our previous simulations that the choice of auxiliary features is crucial for the classification errors. For benchmark purposes, the average classification errors using p-LDA on the MDS-embedded data are $48.40\%$ for GE and $56.65\%$ for GF, which are slightly better than the average LDA errors using PCA projected data but worse than the average LDA errors using multiple datasets and the best GCCA/CCA projections at $d=20$ in this experiment.

Unfortunately, one cannot easily check the joint distribution by Definition~\ref{XYZ} like in the simulation part, because the optimal projection $A^{*}$ is unknown for the Wikipedia datasets. Therefore in a real-world application, one must be cautious in adding a new dataset and/or choosing the best dimension. Both of these are difficult model selection problems in practice, which can be addressed by cross-validation as in this experiment. Still, the interpretation after Definition~\ref{XYZ} is useful from a qualitative perspective. On one hand, the graph dissimilarities GE and GF are of questionable value because they depend on the Internet links, which may be erroneous. On the other hand, the text dissimilarities TE and TF are much more faithful because they are extracted from the document contents, thus more likely to have commonality in certain ``signal'' dimensions. Therefore it is reasonable to believe that choosing a text dissimilarity is better than choosing a graph dissimilarity, which explains why the best $A_{2}$ and $A_{3}$ do not choose any graph dissimilarity as the auxiliary variable and why $A_{4}$ performs worse.

\section{Proofs}
\label{append}
\subsection{Proof of Theorem~\ref{main} when $K=2$ and $r=1$}
\begin{proof} 
We consider $K=2$ and $r=1$ here (and generalize in the next proof), so the number of classes is two and the GCCA criterion is the sum of correlations.  

If a projection $A$ represents the same subspace as the optimal projection $A^{*}$ (i.e., $AA^{'}=A^{*}A^{*'}$), then $A$ is optimal for classification such that $L_{A}=L_{A^{*}}$. For most parts it suffices to assume that $A^{*}$ is unique (in the sense of representing the same subspace), which is justified towards the end of the proof. 

In addition to the uniqueness of $A^{*}$, we also assume that $H_{X}, H_{Z_{s}}, \Sigma_{Z_{s}}$ are all identity matrices for $s=1,2$. This is also justified later, as we will show the theorem is invariant under proper transformations. Further, the expectations $E(X)$ and $E(Z_{s})$ are treated as zeros throughout all proofs because the GCCA/CCA projections and the classification task are not affected.

Under the above assumptions, we have the following: the optimal projection $A^{*}$ is spanned by the first $d$ coordinate axes; any potential projection $A \in \mathcal{A}$ must be orthonormal and equivalent to an orthogonal projection onto a dimension $d$ linear subspace; and the GCCA/CCA projections $A_{s+1}$ are optimal if and only if $A_{s+1}A_{s+1}^{'}=A^{*}A^{*'}$. 

Because all the pre-multiplication matrices are assumed to be identity matrices, together with conditions (1) and (2) in Definition~\ref{XYZ} we have the covariance matrices
\begin{align*}
\Sigma_{XZ_{1}} &= \left[ {\begin{array}{cc}
 p q_{11}E(U_{1}U_{1}^{'})+(1-p)q_{12}E(U_{2}U_{2}^{'}) & p E(U_{1}W_{1}^{'})+ (1-p) E(U_{2}W_{1}^{'})  \\
 \\
 p q_{11}E(WU_{1}^{'})+ (1-p) q_{12}E(WU_{2}^{'}) & E(WW_{1}^{'})  \\ \end{array} } \right]\\
&= \left[ {\begin{array}{cc}
 (p q_{11}+(1-p)q_{12})I_{d \times d} & 0\\
 \\
 0 & E(WW_{1}^{'})  \\ \end{array} } \right],
\end{align*}
\begin{align*}
\Sigma_{XZ_{2}} &= \left[ {\begin{array}{cc}
 p q_{21}E(U_{1}U_{1}^{'})+(1-p)q_{22}E(U_{2}U_{2}^{'}) & p E(U_{1}W_{2}^{'})+ (1-p) E(U_{2}W_{2}^{'})  \\
 \\
 p q_{21}E(WU_{1}^{'})+ (1-p) q_{22}E(WU_{2}^{'}) & E(WW_{2}^{'})  \\ \end{array} } \right]\\
&=\left[ {\begin{array}{cc}
 (p q_{21}+(1-p)q_{22})I_{d \times d} & 0  \\
 \\
 0 & E(WW_{2}^{'})  \\ \end{array} } \right],
\end{align*}
where we denote $p_{1}=p$ and $p_{2}=1-p$ in case of two classes. 

To derive the CCA projection $A_{2}=A_{2}(X,Z_{1})$, the two $m \times d$ orthonormal matrices $A_{2}$ and $B_{2}$ shall maximize the singular values of $A_{2}^{'}\Sigma_{XZ_{1}}B_{2}$ (we take $B_{2}=[b_{1},\ldots,b_{d}]$ as in Equation~(\ref{ccaCond}), similarly to how we define $A_{2}$) \cite{HornJohnsonBook}. Because $A^{*}$ represents the dimension $d$ subspace spanned by the first $d$ coordinate axes, $A_{2}(X,Z_{1})$ is optimal if and only if $A_{2}$ consists of the first $d$ left singular vectors of $\Sigma_{XZ_{1}}$. Due to the form of $\Sigma_{XZ_{1}}$, in this case $B_{2}$ must consist of the first $d$ right singular vectors and the respective correlations are maximized to the decreasingly ordered singular values of the $d \times d$ leading principal sub-matrix of $\Sigma_{XZ_{1}}$. Therefore $A_{2}A_{2}^{'}=A^{*}A^{*'}$ if and only if $A_{2}$ is spanned by the first $d$ coordinate axes, or equivalently the largest $d$ singular values of $\Sigma_{XZ_{1}}$ all come from the $d \times d$ leading principal sub-matrix. 

Putting into inequalities, the CCA projections $A_{2}(X,Z_{s})$ are optimal if and only if
\begin{equation}
\label{cond1}
h_{s}=p q_{s1}+(1-p)q_{s2} - \sigma_{1}(E(WW_{s}^{'}))  > 0.
\end{equation}
When either CCA projections is not optimal, at least one $h_{s}$ is non-positive and represents the ``singular value loss" of using CCA. 

To derive the GCCA projection $A_{3}$ based on $(X,Z_{1},Z_{2})$, the covariance matrix between $Z_{1}$ and $Z_{2}$ also comes into play:
\begin{align*}
\Sigma_{Z_{1}Z_{2}} &=\left[ {\begin{array}{cc}
 p q_{11}q_{21}E(U_{1}U_{1}^{'})+ (1-p) q_{12}q_{22}E(U_{2}U_{2}^{'}) & p q_{11}E(U_{1}W_{2}^{'})+ (1-p) q_{12}E(U_{2}W_{2}^{'})  \\
 \\
 p q_{21}E(W_{1}U_{1}^{'})+ (1-p) q_{22}E(W_{1}U_{2}^{'}) & E(W_{1}W_{2}^{'})  \\ \end{array} } \right]\\
&=\left[ {\begin{array}{cc}
 (p q_{11}q_{21}+ (1-p) q_{12}q_{22})I_{d \times d} & 0  \\
 \\
 0 & E(W_{1}W_{2}^{'})  \\ \end{array} } \right].
\end{align*}
Argued in a similar manner, the GCCA projection is optimal if and only if $A_{3}$ is spanned by the first $d$ coordinate axes. The necessary and sufficient condition for that is
\begin{equation}
\label{cond2}
h+ h_{1}+h_{2}  > 0,\\
\end{equation}
where we define $h=p q_{11}q_{21}+ (1-p) q_{12}q_{22}  -\sigma_{1}(E(W_{1}W_{2}^{'}))$. In words, if both the CCA projections are already optimal, it is sufficient that the largest $d$ singular values of $\Sigma_{Z_{1}Z_{2}}$ all come from the $d \times d$ leading principal sub-matrix; else if either CCA projections is not optimal, the ``singular value gain" from $\Sigma_{Z_{1}Z_{2}}$ has to compensate the possible ``singular value loss" from $\Sigma_{XZ_{1}}$ and $\Sigma_{XZ_{2}}$ in order for the GCCA projection to be optimal. 

An important step is to prove that if $h_{s} \geq 0$ for $s=1,2$, then $h > 0$. This is true because
\begin{align*}
h &= p q_{11}q_{21}+ (1-p) q_{12}q_{22} -  \sigma_{1}(E(W_{1}W_{2}^{'}))  \\
&\geq p q_{11}q_{21}+ (1-p) q_{12}q_{22} -  \sigma_{1}(E(WW_{1}^{'}))\sigma_{1}(E(WW_{2}^{'}))  \\
&\geq p q_{11}q_{21}+ (1-p) q_{12}q_{22} - (p q_{11}+(1-p)q_{12})(p q_{21}+(1-p)q_{22}) \\
&= p(1-p)(q_{11}-q_{12})(q_{21}-q_{22}) \\
& >  0,
\end{align*}
where the first inequality uses condition (3) in Definition~\ref{XYZ}, the second inequality is by the fact that $h_{s} \geq 0$, and the last inequality uses condition (4). 

By the above derivation, if both CCA projections are optimal such that $h_{s} > 0$ for $s=1,2$, then Inequality~(\ref{cond2}) automatically holds and the GCCA projection $A_{3}$ is also optimal. This shows that any $F_{3} \in \Omega_{3}^{*}$ satisfying Inequality~(\ref{cond1}) for $s=1,2$ is an element of the subset $\{F_{3}\in \Omega_{3}^{*}| \max{\{L_{A_{2}}\}}=L_{A_{3}}=L_{A^{*}}\}$.

Next we show there exists $F_{3} \in \Omega_{3}^{*}$ such that Inequality~(\ref{cond2}) holds while Inequality~(\ref{cond1}) fails for at least one $s$. The trivial example is that: if $h_{1} = h_{2} = 0$, then the GCCA projection is optimal. Furthermore, fixing $h$, $p$ and all the $q_{sk}$, the left-hand side of Inequality~(\ref{cond2}) is clearly continuous with respect to $\sigma_{1}(E(WW_{s}^{'}))$ for each $s$. This means $\sigma_{1}(E(WW_{s}^{'}))$ can be increased such that $h_{s}<0$ (and condition (3) in Definition~\ref{XYZ} will not be violated) while Inequality~(\ref{cond2}) still holds. So there also exists $F_{3}$ such that the GCCA projection is optimal when $h_{s} <0$. Thus $\exists F_{3} \in \{F_{3}\in \Omega_{3}^{*}| \max{\{L_{A_{2}}\}} > L_{A_{3}}=L_{A^{*}}\}$. 

Therefore, when $A^{*}$ is unique and $H_{X},H_{Z_{s}}, \Sigma_{Z_{s}}$ are all identity matrices, we proved that: for any given $F_{3} \in \Omega_{3}^{*}$, if the CCA projections are optimal, so are the GCCA projections; if the CCA projections are not optimal (Inequality~(\ref{cond1}) is not satisfied for at least one $s$), the GCCA projection may be optimal (depending on whether the covariance structure satisfies Inequality~(\ref{cond2})). Equivalently, we demonstrate that the similarity definition is sufficient for GCCA to improve CCA. Note that the step that ensures $h>0$ when $h_{s} \geq 0$ will be used again.

Next we show that the result so far is invariant under any $H_{X}, H_{Z_{s}}, \Sigma_{Z_{s}}$ that satisfy Definition~\ref{XYZ}. Take CCA on $(X, Z_{1})$ for an example: by Equation~(\ref{X}) and Equation~(\ref{YZZ}) we have $\Sigma_{\tilde{X}}=H_{X}\Sigma_{X}H_{X}^{'}=I$ and $\Sigma_{\tilde{Z}_{1}}=H_{Z_{1}}\Sigma_{Z_{1}}H_{Z_{1}}^{'}$; also by eigendecomposition there exists $m_{1} \times m_{1}$ matrix $V$ s.t.\ $\Sigma_{\tilde{Z}_{1}}=V^{'}V$. Then $\Sigma_{X} = H_{X}^{-1}H_{X}^{-1'}$ and $\Sigma_{Z_{1}}=H_{Z_{1}}^{-1}V^{'}(H_{Z_{1}}^{-1}V^{'})^{'}$, and the CCA formulation~(\ref{ccaCond}) is equivalent to 
\begin{align*}
&\rho_{\{a_{i}^{'}X,b_{i}^{'}Z_{1}\}}=\frac{(H_{X}^{-1'}a_{i})^{'}H_{X}^{'}\Sigma_{XZ_{1}}H_{Z_{1}}^{'}V^{-1}(VH_{Z_{1}}^{-1'})b_{i}}{\sqrt{(H_{X}^{-1'}a_{i})^{'}H_{X}^{-1'}a_{i}}\sqrt{(VH_{Z_{1}}^{-1'}b_{i})^{'}VH_{Z_{1}}^{-1'}b_{i}}},\\
\mbox{ subject to }&\rho_{\{a_{i}^{'}X,a_{j}^{'}X\}} = \frac{(H_{X}^{-1'}a_{i})^{'}H_{X}^{-1'}a_{j}}{\sqrt{(H_{X}^{-1'}a_{i})^{'}H_{X}^{-1'}a_{i}}\sqrt{(H_{X}^{-1'}a_{j})^{'}H_{X}^{-1'}a_{j}}}=0 \\
\mbox{ and } &\rho_{\{b_{i}^{'}Z_{1},b_{j}^{'}Z_{1}\}}=\frac{(VH_{Z_{1}}^{-1'}b_{i})^{'}VH_{Z_{1}}^{-1'}b_{j}}{\sqrt{(VH_{Z_{1}}^{-1'}b_{i})^{'}VH_{Z_{1}}^{-1'}b_{i}}\sqrt{(VH_{Z_{1}}^{-1'}b_{j})^{'}VH_{Z_{1}}^{-1'}b_{j}}}=0,
\end{align*}
where $V^{-1}$ is defined as the unique Moore-Penrose pseudo inverse if $\Sigma_{\tilde{Z}_{1}}$ is singular. Hence it is equivalent to consider the projections $H_{X}^{-1'}A_{2}$ and $VH_{Z_{1}}^{-1'}B_{2}$ on $(\tilde{X}, V^{-1'}\tilde{Z}_{1})$ (both $\tilde{X}$ and $V^{-1'}\tilde{Z}_{1}$ are of identity variance) with covariance $H_{X}^{'}\Sigma_{XZ_{1}}H_{Z_{1}}^{'}V^{-1}$, instead of the projections $A_{2}$ and $B_{2}$ on $(X, Z_{1})$. The same holds for the GCCA formulation~(\ref{gccaCond}). Furthermore, the classification task remains the same because the projected feature $A^{'}X=(H_{X}^{-1'}A)^{'}H_{X}X$ is invariant under the full-rank transformation $H_{X}$. Therefore the optimal projection $A^{*}$ and the GCCA/CCA projections $A_{s+1}$ are all equivalent to the identity variance case up to $H_{X}$, and the result is clearly invariant. 

At last we justify the case when $A^{*}$ is not unique, which means there exists $A^{*}$ that is spanned by the first $d$ coordinate axes under different transformation matrices. Because the conditions in Definition~\ref{XYZ} are required to be satisfied for all $A^{*}$, in most cases the CCA optimality is still equivalent to Inequality~(\ref{cond1}), i.e., CCA is optimal if and only if Inequality~(\ref{cond1}) is satisfied for at least one $A^{*}$ after proper transformations for each $A^{*}$. The same holds for the GCCA optimality (Inequality~(\ref{cond2})), and we can still conclude that GCCA improves CCA following the same steps. However, a special case should be taken into consideration, and we take the CCA projection $A_{2}(X,Z_{1})$ for an illustration: Suppose the singular vector $\sigma_{1}(E(WW_{s}^{'}))$ corresponds to is the $(d+1)$th coordinate axes and $\sigma_{1}(E(WW_{s}^{'})) > \sigma_{2}(E(WW_{s}^{'}))$. Then $A_{2}(X,Z_{1})$ can be chosen to represent any dimension $d$ subspace of the space spanned by the first $(d+1)$ coordinate axes, and the degrees of freedom is $(d+1)d-\frac{d^2+d}{2}$ (the degrees of freedom may increase if there are repeating singular values). Now, if $A^{*}$ happens to have the same degrees of freedom in the space spanned by the first $(d+1)$ coordinate axes, then $A_{2}(X,Z_{1})$ is optimal if and only if $h_{1} \geq 0$ (rather than $h_{1} >0$) because any arbitrary choice of $A_{2}$ is optimal. Similar phenomenon applies for $A_{s+1}$, in which case Inequality~(\ref{cond1}) and Inequality~(\ref{cond2}) should be adjusted to include equalities. However, in this case we still have $h+h_{1}+h_{2} >0$ when the CCA projections are optimal, which is still sufficient (but may not be necessary) for GCCA to be optimal. Therefore, GCCA still improves CCA in case of non-unique $A^{*}$, and the justification is done. \qed
\end{proof}

\subsection{Proof of Theorem~\ref{main} for any $K \geq 2$ and $r \geq 1$}
\begin{proof}
Now we generalize the result to arbitrary $K \geq 2$ (multi-class) and any $r \geq 1$ (the GCCA criterion). Without loss of generality, we assume that $A^{*}$ is unique and $H_{X}, H_{Z_{s}}, \Sigma_{Z_{s}}$ are all identity matrices. 

Let us treat the case that $r=1$ first. Using the setting in Equation~(\ref{YZZ}) and argue similarly as before, GCCA improves CCA if and only if
\begin{equation}
\label{multiclass}
h=\sum_{k=1}^{K} p_{k}q_{1k}q_{2k} -  \sigma_{1}(E(W_{1}W_{2}^{'}))  > 0
\end{equation}
is true when $h_{s}= \sum_{k=1}^{K} p_{k}q_{sk} - \sigma_{1}(E(WW_{s}^{'})) \geq 0$ for $s=1,2$.

This is true because
\begin{align}
\label{hmul}
h &= \sum_{k=1}^{K} p_{k}q_{1k}q_{2k}-  \sigma_{1}(E(W_{1}W_{2}^{'})) \nonumber \\
&\geq \sum_{k=1}^{K} p_{k}q_{1k}q_{2k}-  \sigma_{1}(E(WW_{1}^{'}))\sigma_{1}(E(WW_{2}^{'})) \nonumber \\
&\geq \sum_{k=1}^{K} p_{k}q_{1k}q_{2k} - (\sum_{k=1}^{K} p_{k}q_{1k}) (\sum_{k=1}^{K} p_{k}q_{2k}) \nonumber \\
&= \sum_{1 \leq k_{1} < k_{2} \leq K} p_{k_{1}}p_{k_{2}}(q_{1k_{1}}-q_{1k_{2}})(q_{2k_{1}}-q_{2k_{2}}) \\
&> 0, \nonumber
\end{align}
where the first inequality follows from conditions (3), the second inequality follows from $h_{s} \geq 0$, the next equality follows from simple algebra, and the last inequality follows from condition (4).  

As to the GCCA criterion with $r \geq 1$, GCCA improves CCA if and only if
\begin{align*}
(\sum_{k=1}^{K}p_{k}q_{1k}q_{2k})^{r} -  \sigma_{1}^{r}(E(W_{1}W_{2}^{'}))  > 0
\end{align*}
is true when $h_{s} \geq 0$. Clearly this inequality holds if and only if it holds for $r=1$, which is Inequality~(\ref{multiclass}). Hence it is true and GCCA improves CCA in the similar family for any $r \geq 1$. 

Thus Theorem~\ref{main} is proved for any number of classes and any GCCA criterion with $r \geq 1$. \qed
\end{proof}

\subsection{Proof of Corollary~\ref{main2} and Corollary~\ref{main3}}

\begin{proof}
Without loss of generality, we carry out the proof assuming $A^{*}$ is unique, $H_{X}, H_{Z_{s}}, \Sigma_{Z_{s}}$ are all identity matrices, and $K=2$ and $r=1$. 

There are $S$ auxiliary features in total, and thus $\dbinom{S}{S^{'}}$ choices of auxiliary features for $A_{S^{'}+1}$. We define $h_{s}=p q_{s1}+(1-p)q_{s2}-\sigma_{1}(E(WW_{s}^{'}))$ and $h_{st}=p q_{s1}q_{t1}+ (1-p) q_{s2}q_{t2}  - \sigma_{1}(E(W_{s}W_{t}^{'}))$ for any $s$ and $t$ satisfying $S \geq s,t \geq 1$, where $h_{st}$ is a generalization of $h$ in the proof of Theorem~\ref{main}. 

Then the GCCA projection $A_{S^{'}+1}$ using the first $S^{'}$ auxiliary features is optimal if and only if
\begin{equation}
\label{cond3}
\sum_{1 \leq s<t \leq S^{'}} h_{st}+ \sum_{s=1}^{S^{'}}h_{s} > 0.
\end{equation}
This is a generalization of Inequality~(\ref{cond2}), because there are $S^{'}$ possible ``singular value loss" caused by $\Sigma_{XZ_{s}}$ and $\frac{S^{'}(S^{'}-1)}{2}$ additional cross-covariance terms $\Sigma_{Z_{s}Z_{t}}$ between the auxiliary features. Note that for any other $A_{S^{'}+1} \in \mathcal{A}_{S^{'}+1}$ with a different choice of auxiliary features, we can still use Inequality~(\ref{cond3}) for the optimality by switching the first $S^{'}$ auxiliary features with the chosen $S^{'}$ auxiliary features.

All the CCA projections are optimal if and only if $h_{s} >0$ for all $s=1,\ldots,S$. This implies that $h_{st}>0$ is always true for any $1 \leq s<t \leq S$, and Inequality~(\ref{cond3}) holds for any $A_{S^{'}+1} \in \mathcal{A}_{S^{'}+1}$ with $S \geq S^{'} \geq 2$. Therefore the set of GCCA projections $\mathcal{A}_{S^{'}+1}$ always improves the set of CCA projections $\mathcal{A}_{2}$, and Corollary~\ref{main2} is proved.

To prove Corollary~\ref{main3}, we use the simplifying condition (4'). Then Inequality~(\ref{cond3}) simplifies to $\frac{S^{'}-1}{2} h_{12} + h_{1} >0$, because $h_{st}$ are the same for all $1 \leq s,t \leq S^{'}$ and so are $h_{s}$. We need to show that if $A_{S^{'}}$ are optimal for certain $F_{S+1}$, so is $A_{S^{'}+1}$. (note that the choice of auxiliary features no longer matters because they follow the same distribution, which means all the elements in $\mathcal{A}_{S^{'}+1}$ represent the same subspace.)

When $S^{'}=2$, it is a special case of Theorem~\ref{main} because any $F_{S+1}$ satisfying condition (4') also satisfies condition (4). Clearly $A_{2}$ is optimal if and only if $h_{1}=h_{2} >0$, which implies $h_{12} >0$. So Inequality~(\ref{cond3}) holds and $A_{3}$ is also optimal.

When $S^{'}=3$, $A_{3}$ is optimal if and only if $h_{12}+h_{1}>0$. In this case if $h_{1} >0$, then we have $h_{12} > 0$; if $h_{1}<0$, then $h_{12} >0$ must be true in order for $A_{3}$ to be optimal. In any case, $\frac{3}{2}h_{12}+h_{1} >0$ is true and $A_{4}$ is optimal.

Therefore, the optimality of $A_{3}$ implies the optimality of $A_{4}$. By induction, for any $S \geq S^{'} \geq 2$, the optimality of $\mathcal{A}_{S^{'}}$ implies the optimality of $A_{S^{'}+1}$ under the simplifying condition (4'), and Corollary~\ref{main3} is proved. Note that the corollary is not true under the original condition (4), and one can easily make up a counter-example by checking Inequality~(\ref{cond3}). \qed 
\end{proof}

\subsection{Comments}
We conclude the proof section by considering the term $h=\sum_{k=1}^{K} p_{k}q_{1k}q_{2k} -  \sigma_{1}(E(W_{1}W_{2}^{'})) $ in Equation~\ref{multiclass} for the case of two auxiliary features, which offers additional insights for Definition~\ref{XYZ} of the similar family and is potentially useful for model selection.

Firstly, the equation offers a relaxation of condition (4) in the similar family: instead of $(q_{sk_{1}}-q_{sk_{2}})(q_{tk_{1}}-q_{tk_{2}}) > 0$ for all $1 \leq s < t \leq S$ and $k_{1},k_{2}=1,\ldots,K$, we can replace it by either $h>0$ or $\sum_{1 \leq k_{1} < k_{2} \leq K} p_{k_{1}}p_{k_{2}}(q_{1k_{1}}-q_{1k_{2}})(q_{2k_{1}}-q_{2k_{2}}) > 0$ (by Equation~\ref{hmul}), which is more difficult to interpret than the original condition but less restrictive.

Secondly, the improvement of GCCA over CCA depends almost solely on the magnitude of $h$. The larger the $h$, the more likely that GCCA may be optimal even if CCA is not. Towards this direction, the magnitude of $q_{sk}$ plays an important role: for fixed $E(W_{1}W_{2}^{'})$, assuming all coefficients non-negative, $h$ increases with $q_{sk}$ and GCCA projection is potentially more superior. 

Finally, the above observation may be useful for the choice of auxiliary variables and the projecting dimension without using cross-validation. Other things being equal, an auxiliary variable with larger $h$ or $q_{sk}$ is more favorable, as is a projection dimension with larger $h$ or $q_{sk}$; thus it is reasonable to choose an auxiliary variable and/or a projection dimension with a more significant ``signal'' part (where $U_{k}$ lives) for later inference, which agrees with intuition. Numerically, within the similar family this observation is useful for model selection purposes (choose the auxiliary feature and/or the projection dimension with the largest $h$ using greedy algorithms, among all available auxiliary features and all possible dimensions); but out of the similar family definition, whether a modified version of $h$ can serve the model selection purpose or not requires further investigation.

\section*{Acknowledgments}
\addcontentsline{toc}{section}{Acknowledgment}
This work was partially supported by National Security Science and Engineering Faculty Fellowship (NSSEFF),
 Johns Hopkins University Human Language Technology Center of Excellence (JHU HLT COE), and the
 XDATA program of the Defense Advanced Research Projects Agency (DARPA) administered through Air Force Research Laboratory
 contract FA8750-12-2-0303.

The authors would like to thank the reviewers for their insightful and valuable suggestions in improving the exposition of the paper.

\bibliographystyle{ieeetr}
\bibliography{references}

\end{document}